\documentclass{article}

\usepackage[nonatbib, preprint]{neurips_2021}

\usepackage[utf8]{inputenc} 
\usepackage[T1]{fontenc}    
\usepackage{hyperref}       
\usepackage{url}            
\usepackage{booktabs}       
\usepackage{amsfonts}       
\usepackage{nicefrac}       
\usepackage{microtype}      
\usepackage{xcolor}         
\usepackage{graphicx}
\usepackage{lscape}
\usepackage{adjustbox}
\usepackage{amssymb}
\usepackage{pifont}
\usepackage{float}

\title{Generative Pretraining for Paraphrase Evaluation}

\author{%
  Jack Weston \\
  Novoic \\
  \texttt{jack@novoic.com} \\
  \And
  Rapha\"{e}l Lenain \\
  Novoic \\
  \texttt{raphael@novoic.com} \\
  \And
  Udeepa Meepegama \\
  Novoic \\
  \texttt{udeepa@novoic.com} \\
  \And
  Emil Fristed \\
  Novoic \\
  \texttt{emil@novoic.com} \\
}

\begin{document}

\maketitle

\begin{abstract}
We introduce ParaBLEU, a paraphrase representation learning model and evaluation metric for text generation. Unlike previous approaches, ParaBLEU learns to understand paraphrasis using generative conditioning as a pretraining objective. ParaBLEU correlates more strongly with human judgements than existing metrics, obtaining new state-of-the-art results on the 2017 WMT Metrics Shared Task. We show that our model is robust to data scarcity, exceeding previous state-of-the-art performance using only $50\%$ of the available training data and surpassing BLEU, ROUGE and METEOR with only $40$ labelled examples. Finally, we demonstrate that ParaBLEU can be used to conditionally generate novel paraphrases from a single demonstration, which we use to confirm our hypothesis that it learns abstract, generalized paraphrase representations.
\end{abstract}

\section{Introduction}\label{section:introduction}
Representing the relationship between two pieces of text, be it through a simple algorithm or a deep neural network, has a long history and diverse use-cases that include the evaluation of text generation models \cite{wiseman2017challenges, van2019best} and the clinical evaluation of human speech \cite{johnson2003discourse, weintraub2018version}. One of the earliest examples of such a representation is the Levenshtein distance \cite{levenshtein1966binary}, which describes the number of character-level edits required to transform one piece of text into another. This metric now forms part of a wider family of edit-distance-based metrics that includes the word error rate (WER) and the translation error rate (TER) \cite{och2003minimum}. Other algorithms, such as ROUGE \cite{lin2004rouge}, METEOR \cite{banerjee2005meteor} and the widely used BLEU metric \cite{papineni2002bleu}, perform exact or approximate $n$-gram matching between the two texts.

These low-level approaches bear little resemblance to the human process of comparing two texts, which benefits from a deep prior understanding of the semantic and syntactic symmetries of language \cite{novikova2017we}. For example, pairs like ``she was no ordinary burglar'' and ``she was an ordinary burglar'' are close in edit-distance-space but semantically disparate. The goal of an automatic text evaluation metric is typically to be a good proxy for human judgements, which is clearly task-dependent. More recently, neural approaches have begun to close the gap between automatic and human judgements of semantic text similarity using Transformer-based language models such as BERT \cite{zhang2019bertscore, sellam2020bleurt}. They aim to leverage the transferable knowledge gained by the model during pretraining on large text corpora. The relationship between two texts is similarly modelled, albeit implicitly, by sequence-to-sequence models such as BART \cite{lewis2019bart} and T5 \cite{raffel2019exploring}. We consider paraphrase evaluation and paraphrase generation to be two instances of \emph{paraphrase representation learning}.

Linguistically, a paraphrase is a restatement that preserves essential meaning, with arbitrary levels of literality, fidelity and completeness. In practice, what qualifies as a good paraphrase is context-specific. One motivation for considering paraphrase evaluation as a representation learning problem is the varied nature of paraphrase evaluation tasks, which may have an emphasis on semantic equivalence (e.g. PAWS \cite{zhang2019paws} and MRPC \cite{dolan2005automatically}), logical entailment versus contradiction (e.g. MultiNLI \cite{williams2017broad} and SNLI \cite{bowman2015snli}), and the acceptability of the generated text (e.g. the WMT Metrics Shared Task \cite{bojar-etal-2017-results}). Considering even broader applications such as clinical speech analysis further motivates learning generalized paraphrase representations. 

In this paper, we introduce ParaBLEU, a paraphrase representation learning model that predicts a conditioning factor for sequence-to-sequence paraphrase generation as one of its pretraining objectives, inspired by style transfer in text-to-speech \cite{skerry2018towards} and text generation systems \cite{yang2018unsupervised, lample2018multiple}. ParaBLEU addresses the primary issue with neural paraphrase evaluation models to date: the selection of a sufficiently generalized pretraining objective that primes the model for strong performance on downstream paraphrase evaluation tasks when data is scarce. Previous state-of-the-art neural models have either used a broad multi-task learning approach or eschewed additional pretraining altogether. The former case may encourage the model to learn the biases of inferior or inappropriate metrics, while the latter leaves room for optimization. Non-neural models, such as BLEU, TER, ROUGE and BERTScore \cite{zhang2019bertscore}, benefit from requiring no training data not being subject to domain shift but cannot, however, learn to exploit task-specific nuances of what defines `good' paraphrasing.

We evaluate ParaBLEU's ability to predict human judgements of paraphrases using the English subset of the 2017 WMT Metrics Shared Task. A useful neural text similarity metric should be robust to data scarcity, so we assess performance as a function of the fine-tuning dataset size. Finally, using the ParaBLEU pretraining model as a paraphrase generation system, we explore our hypothesis that the model reasons in high-level paraphrastic concepts rather than low-level edits through an explainability study, and demonstrate that ParaBLEU can operate as a conditional paraphrase generation model.

\section{Approach}
\label{sec:approach}
In this section, we describe and justify the set of inductive biases we build into ParaBLEU, along with a description of the model architecture and pretraining/fine-tuning strategy. We consider a reference text $x$ and a candidate text $\hat x$. We wish to learn a function $f : f(x, \hat x) \rightarrow y$, where $y \in \mathbb{R^N}$ is a single- or multi-dimensional paraphrase representation, which could be a scalar score. 

\subsection{Inductive biases}\label{section:inductive_biases}
Our approach begins by decomposing paraphrase representation learning into three overlapping factors:
\begin{enumerate}
    \item \textbf{Edit-space representation learning: } Building a representation of high-level syntactic and semantic differences between $x$ and $\hat x$, contrasted with the low-level pseudo-syntactic/-semantic operations considered by edit-distance-based and $n$-gram based metrics.
    \item \textbf{Candidate acceptability judgement: } Evaluating the grammaticality, coherence and naturalness of $\hat x$ in isolation. Perplexity \cite{jelinek1977perplexity} with respect to a given language model is one proxy for this.
    \item \textbf{Semantic equivalence: } Assessing whether $x$ and $\hat x$ convey the same essential meaning precisely, as opposed to merely being semantically similar. This is related to entailment classification tasks and, more broadly, the interaction between language and formal logic.
\end{enumerate}
Exploiting this factorization, we hypothesize that the following inductive biases are beneficial to a paraphrase representation learning model:
\begin{itemize}
    \item \textbf{Using pretrained language models: } All three factors require a general understanding of the semantic and syntactic structures of language, making transfer learning from powerful pretrained language models such as BERT \cite{devlin2018bert} appealing.
    \item \textbf{Non-local attention as bitext alignment: } Factors (1) and (3) require performing context-aware `matching' between $x$ and $\hat x$. This is similar to the statistical method of bitext alignment \cite{tiedemann2011bitext}. Attention mechanisms within a Transformer \cite{vaswani2017attention} are an obvious candidate for learnable context-aware matching, which has precedent in paraphrasing tasks and the next-sentence-prediction objective of the original BERT pretraining. We note that $x$ and $\hat x$ side-by-side violates attention locality, meaning local attention mechanisms, such as those used in T5, may be suboptimal for longer text-pairs.
    \item \textbf{Bottlenecked conditional generation objective: } A key insight is that a strong factor (1) representation $z \in \mathbb{R}^M$ where $h: h(x, \hat x) \rightarrow z$ is one that can condition the sampling of $\hat x$ from $x$ through some generative model $g : g(x\;|\;z) \rightarrow \hat x$. One trivial solution to this is $h(x, \hat x) = \hat x$. To avoid this case, we introduce an information bottleneck on $z$ such that it is advantageous for the model to learn to represent high-level abstractions, which are cheaper than copying $\hat x$ through the bottleneck if they are sufficiently abstract compared with $\hat x$, the bottleneck is sufficiently tight, and the decoder can jointly learn the same abstractions. It is likely advantageous to use a pretrained sequence-to-sequence language model, which can already reason in linguistic concepts. 
    \item \textbf{Masked language modelling objective: } Factor (2) can be addressed by an MLM objective, which alone is sufficient for a neural network to learn a language model \cite{devlin2018bert}. Performing masked language modelling on a reference-candidate pair also encourages the network to use $x$ to help unmask $\hat x$ and vice versa, strengthening the alignment bias useful for factors (1) and (2).
    \item \textbf{Entailment classification objective: } Factor (3) is similar to the classification of whether $x$ logically entails $\hat x$. There are a number of sentence-pair datasets with entailment labels that could be used to construct this loss; see Table \ref{tab:data_table}.
\end{itemize}

\subsection{ParaBLEU}
\begin{figure}[t]
  \centering
  \includegraphics[width=0.8\textwidth]{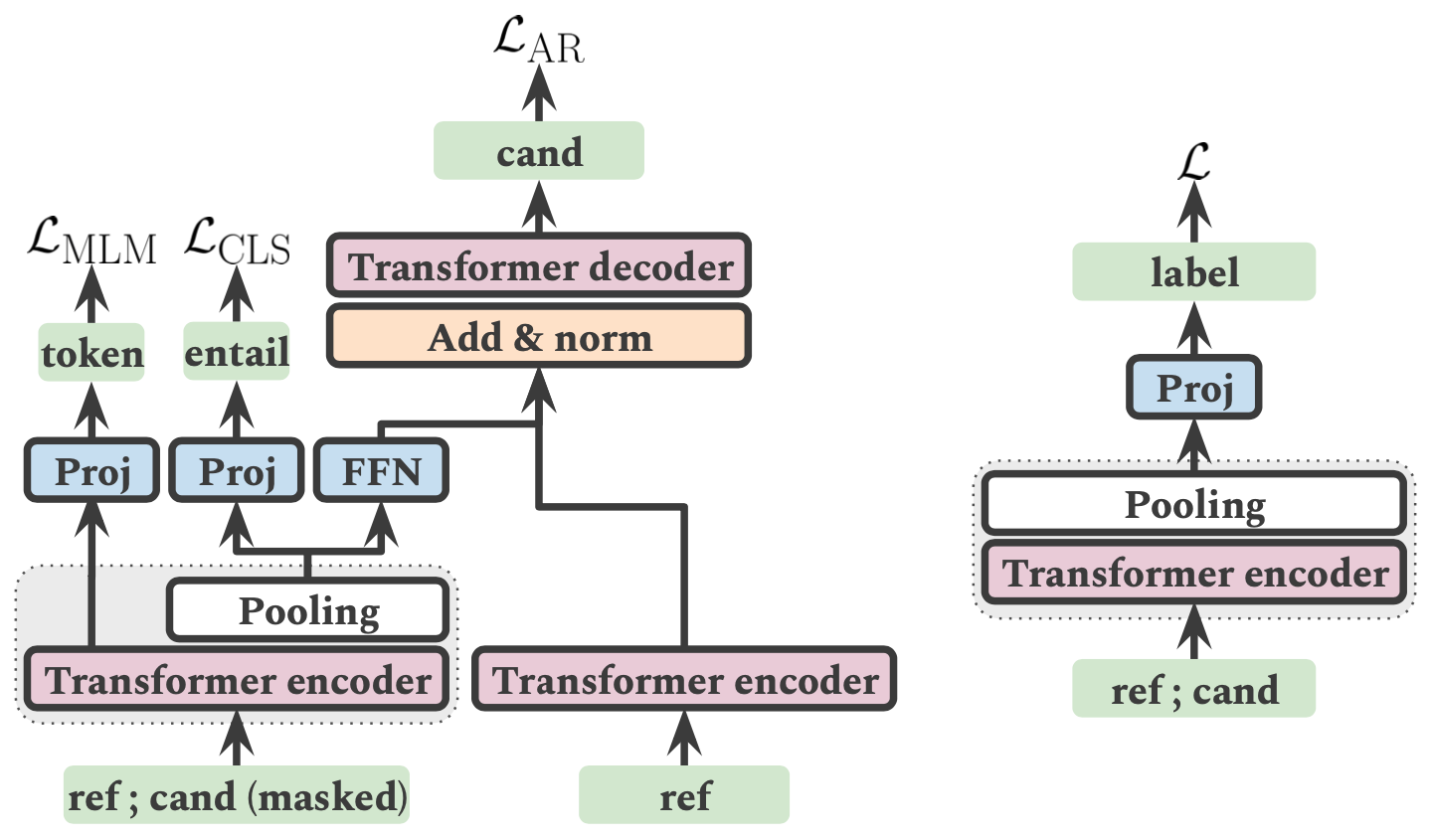}
  \caption{The pretraining (left) and fine-tuning (right) setups for ParaBLEU. `\texttt{ref ; cand}' indicates the canonical method for combining a reference and candidate sentence for a given language model. $\mathcal{L}_{\small\textrm{AR}}$ is an autoregressive causal language modelling loss, $\mathcal{L}_{\small\textrm{MLM}}$ a masked language modelling loss, and $\mathcal{L}_{\small\textrm{CLS}}$ an entailment classification loss. The feedforward network (FFN) includes two affine layers, the middle dimension of which can be used to create a bottleneck (see Section \ref{section:inductive_biases}). Dropout layers and activations are omitted for brevity.}
  \label{figure:architecture}
\end{figure}

Inspired by style transfer in text-to-speech \cite{skerry2018towards} and text generation systems \cite{yang2018unsupervised, lample2018multiple}, we propose the architecture shown in Figure \ref{figure:architecture}. The grey box indicates the Transformer encoder we wish to pretrain, which we refer to as the `edit encoder'. Factorization of the task leads to three complementary objectives: a cross-entropy masked language modelling loss $\mathcal{L}_{\small\textrm{MLM}}$, a binary cross-entropy entailment classification loss $\mathcal{L}_{\small\textrm{CLS}}$ and a cross-entropy autoregressive causal language modelling loss $\mathcal{L}_{\small\textrm{AR}}$. An additional sequence-to-sequence Transformer model is used during pretraining to provide a learning signal. The proposed bottleneck lies within the feedforward network (FFN). The full pretraining loss is given by:
\begin{equation}\label{eq:loss}
    \mathcal{L_\textrm{pre}} :=  \mathcal{L}_{\small\textrm{AR}} + \alpha \cdot \mathcal{L}_{\small\textrm{MLM}} + \beta \cdot \mathcal{L}_{\small\textrm{CLS}},
\end{equation}
where $\alpha$ and $\beta$ are tunable hyperparameters. We probe the importance of each objective in the ablation studies in Section \ref{section:ablations}. At fine-tuning time, the sequence-to-sequence model is discarded and the edit encoder is fine-tuned using a linear projection on top of the pooled output. Throughout this work, our pooling layers simply take the beginning-of-sequence token.

Our architecture places restrictions on valid combinations of pretrained models. We found in practice that using an encoder-only pretrained language model to initialize the edit encoder, and a sequence-to-sequence pretrained language model to initialize the sequence-to-sequence model, works best. This is likely because encoder-only models are encouraged to encode strong representations at the final layer, and these representations have already been directly pretrained with an MLM objective. For technical ease we require that the models have a consistent tokenizer and vocabulary, and that the pretrained checkpoints are available through the HuggingFace \texttt{transformers} library \cite{wolf2019huggingface}. In this paper, we consider the combination RoBERTa \cite{liu2019roberta} + BART, but we note that both multilingual (XLM-R \cite{conneau2019unsupervised} + mBART \cite{liu2020multilingual}) and long (Longformer + Longformer-Encoder-Decoder (LED) \cite{beltagy2020longformer}) combinations exist. We consider both base and large variants, which correspond to RoBERTa$_{\small\textrm{base}}$ and RoBERTa$_{\small\textrm{large}}$. In both cases, we use a BART$_{\small\textrm{base}}$ checkpoint.

\subsection{Related work}\label{section:related_work}
To contextualize this work, we provide a summary of related architectures and describe the ways in which they are similar/dissimilar to our proposed model. BLEURT \cite{sellam2020bleurt} and FSET \cite{kazemnejad2020paraphrase} are the most relevant.

\textbf{BLEURT} is a neural automatic evaluation metric for text generation. Starting from a pretrained BERT model, it is further pretrained to predict a number of pre-existing metrics, such as BLEU, ROUGE and BERTScore. ParaBLEU, by contrast, does not use pre-existing metrics as training objectives, instead using generative conditioning as a more general signal for paraphrase representation learning. \textbf{FSET} is a retrieval-based paraphrase generation system in which a sentence $z$ is paraphrased by first locating a similar reference sentence from a large bank of reference/candidate pairs, then extracting and replaying similar low-level edits on $z$. Common to ParaBLEU and FSET is the use of a Transformer for paraphrase style transfer, with differing architectural details. However, FSET is designed to transpose low-level edits and so requires lexically similar examples; whereas ParaBLEU is explicitly designed to learn high-level, reference-invariant paraphrase representations using a factorized objective. The \textbf{musical style Transformer autoencoder} \cite{choi2020encoding} uses a similar Transformer-based style transfer architecture to conditionally generate new music in controllable styles. Other examples in text-to-speech systems perform style transfer by encoding the prosody of a source sentence into a bottlenecked reference  embedding \cite{skerry2018towards} or disentangled style tokens \cite{wang2018style}.

There is a wealth of recent literature on controllable paraphrase generation and linguistic style transfer, some of which we highlight here. \textbf{T5} leverages a huge text corpus as pretraining for conditional generation using `commands' encoded as text, which includes paraphrastic tasks such as summarization. \textbf{Linguistic style transfer} \cite{yang2018unsupervised, zhao2018language, jin2020deep} work aims to extract the style of a piece of text and map it onto another piece of text without changing its semantic meaning. \textbf{STRAP} \cite{krishna2020reformulating} generates paraphrases in controllable styles by mixing and matching multiple style-specific fine-tuned GPT-2 models. \textbf{REAP} \cite{goyal2020neural} uses a Transformer to syntactically diverse generate paraphrases by including an additional position embedding representing the syntactic tree. \textbf{DNPG} \cite{li2019decomposable} is a paraphrase generation system that uses a cascade of Transformer encoders and decoders to control whether paraphrasing is sentential or phrasal. 

\section{Data}
\begin{figure}[t]
\centering
  \includegraphics[width=0.44\textwidth]{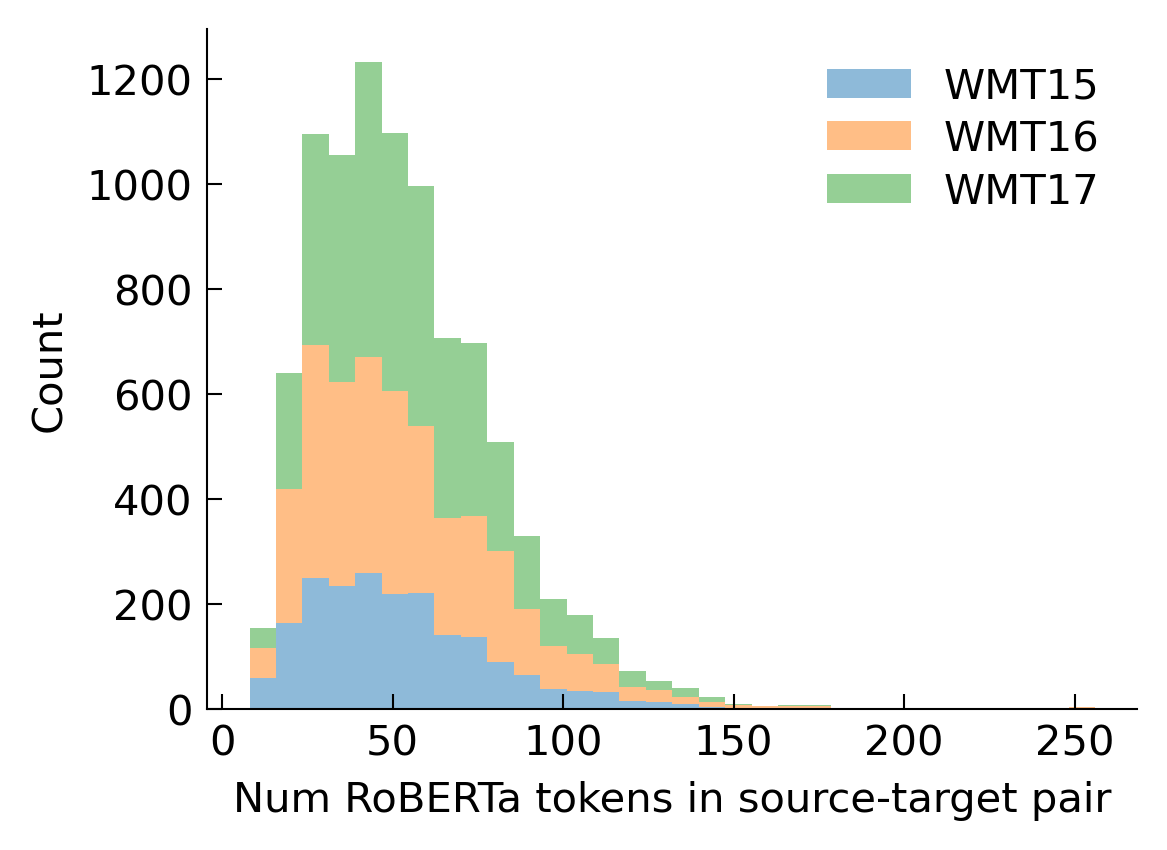}
  \centering
  \includegraphics[width=0.4\textwidth]{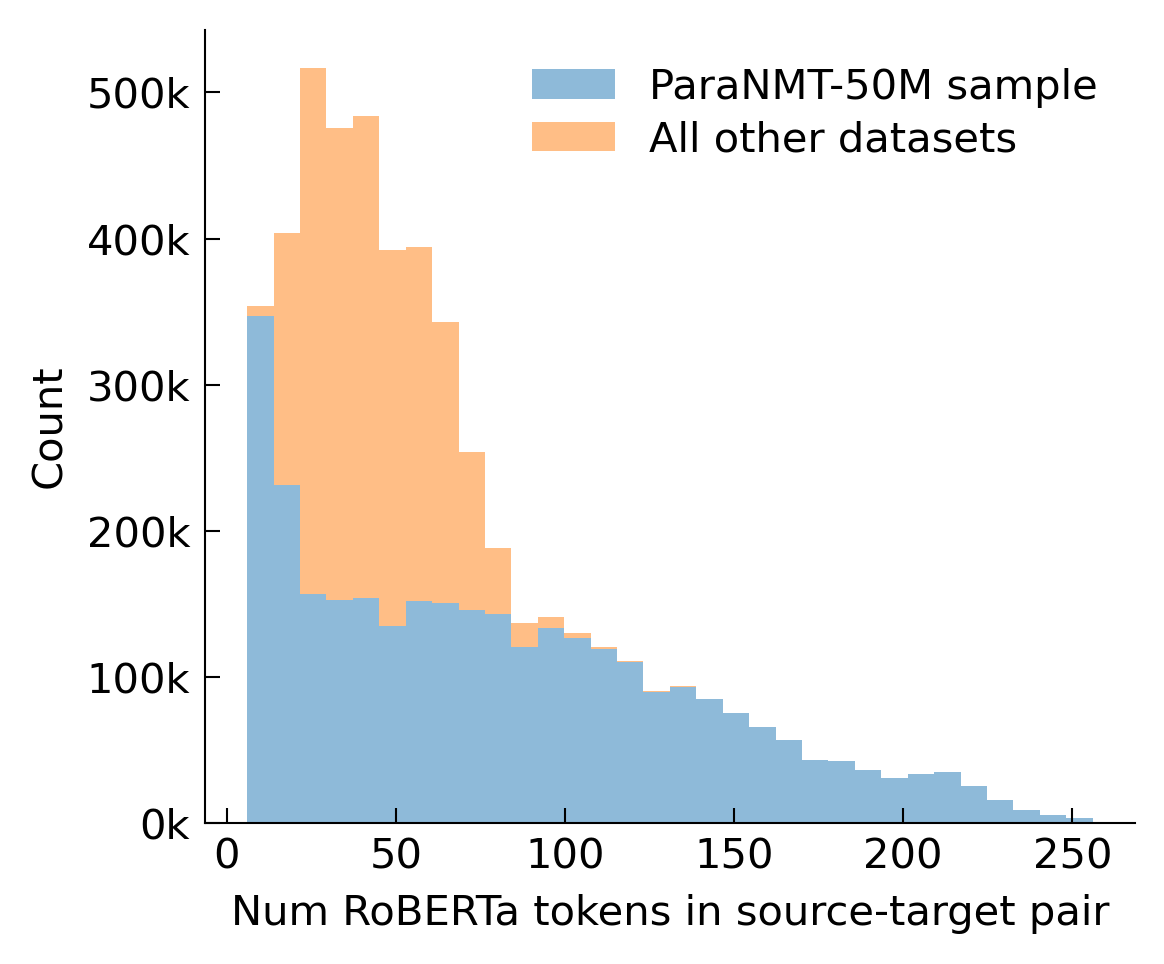}
\caption{Stacked histograms showing the distribution of the number of tokens in the WMT Metrics Shared Task data (left) and ParaCorpus (right).}
\label{fig:data_lens}
\end{figure}

In this section, we describe the pretraining and fine-tuning datasets we use in our studies.

\subsection{WMT Metrics Shared Task}
The WMT Metrics Shared Task is an annual benchmark for automated evaluation metrics for translation systems, where the goal is to predict average human ratings comparing the machine-translated candidate $\hat x$ with human-translated reference $x$, both of which have been translated from the same source sentence. 

We use an identical setup to \cite{sellam2020bleurt} and \cite{zhang2019bertscore}, where we use the subset of data for which the candidate and reference are in English, which we will refer to as the to-English subset. The source, which is unused, can be in any non-English language, the set of which varies from year-to-year. We produce results for the WMT Metrics Shared Task 2017 (WMT17) using the official test, and train on the to-English subsets of WMT15 and WMT16. The training sets contains $5,360$ examples. The distributions of example length in tokens is shown in Figure \ref{fig:data_lens}.

We report the agreement between the metric and the human scores using two related correlation coefficients: absolute Kendall $|\tau|$ and absolute Pearson $|r|$, the latter of which was the official metric of the 2017 task. In our summary results in the main paper, we average these metrics across all source languages but not over reference/candidate language. Full results are provided in Appendix \ref{appendix:full_wmt_results}.

\subsection{ParaCorpus}
\begin{table}[t]
  \caption{ParaCorpus composition.}
  \label{tab:data_table}
  \centering
  \begin{adjustbox}{center}
  \bgroup
  \def\arraystretch{1.5}
  \begin{tabular}{lp{0.25\linewidth}p{0.25\linewidth}ccc}
    \toprule
    Dataset                                             & Subsets included & Nature & Size & Ent. labels & Ref \\
    \midrule
    PAWS         & Wiki-final-train; \newline Wiki-swap-train; \newline Wiki-unlabeled-train; QQP-train    & Entailment sentence pairs with high semantic overlap & 740k & \checkmark & \cite{zhang2019paws} \\
    SNLI        & Train                                                                 & Human-written entailment sentence pairs & $550$k &  \checkmark$\dagger$ & \cite{bowman2015snli} \\
    MultiNLI     & Train                                                                 & Multi-genre entailment sentence pairs & $390$k & \checkmark$\dagger$ & \cite{williams2017broad} \\
    ParaSCI      & ACL-train; arXiv-train                                                & Human-written academic paraphrase pairs & $340$k &  \ding{55} & \cite{dong2021parasci} \\
    ParaNMT-50M  & Random sample (see main text)                                                & Varied paraphrase pairs from machine translation & $3.1$m &  \ding{55} & \cite{wieting2017paranmt} \\
    \midrule
    ParaCorpus       & -                                                                     & - & $5.1$m & Partial & - \\
    \bottomrule
  \end{tabular}
  \egroup
  \end{adjustbox}
\end{table}

In addition to our design choices, we also encourage a robust and generalizable pretraining by using a dataset covers a variety of styles and lengths. We collate a number of paraphrase datasets to create a single pretraining dataset we call ParaCorpus. The composition of the dataset is shown in Table \ref{tab:data_table}, with a total of $\sim5.1$m examples. All examples have reference and candidate texts and around one third additionally have binary entailment labels.  Where the source dataset included three-way labels `entailment'/`contradiction'/`neutral', `entailment' was mapped to $1$ and the others to $0$. A subset of ParaNMT-50M \cite{wieting2017paranmt}, which includes noisier, speech-like examples, was included for two reasons: to add additional stylistic diversity to the dataset, and to increase the population of the dataset with combined token lengths above $128$, which we hypothesize will make the model more robust to the longer examples seen in the WMT datasets. tion tokens lengths is shown in Figure \ref{fig:data_lens}.

\section{Experiments}
\label{sec:experiments}
In this section, we present results on WMT17, benchmarked against the current state-of-the-art approach, along with widely used neural, $n$-gram and edit-distance-based metrics. We study ParaBLEU performance as a function of number of pretraining steps and the size of the fine-tuning dataset. Finally, we perform ablations to test the impact of the inductive biases and resultant architectural decisions described in Section \ref{sec:approach}.

We report results for both ParaBLEU$_{\small\textrm{base}}$, based on RoBERTa$_{\small\textrm{base}}$ ($12$ layers, $768$ hidden units, $12$ heads), and our default model ParaBLEU$_{\small\textrm{large}}$, based on RoBERTa$_{\small\textrm{large}}$ ($24$ layers, $1,024$ hidden units, $16$ heads). Both models are trained near-identically for $4$ epochs on ParaCorpus. Further pretraining details can be found in Appendix \ref{appendix:model_hyperparameters}. For fine-tuning, we use a batch size of $32$, a learning rate of $1$e-$5$ and train for $40$k steps, with a validation set size of $10\%$ (unless otherwise stated). No reference texts are shared between the train and validation sets, following \cite{sellam2020bleurt}. Pretraining ParaBLEU$_{\small\textrm{large}}$ takes $\sim10$h on a $16$ A100 GPU machine. Fine-tuning takes $\sim8$h on a single A100 GPU machine.

\subsection{Results}\label{section:results}
\begin{table}[t]
  \caption{Summary results for WMT17. The metrics reported are absolute Kendall $|\tau|$ and Pearson $|r|$ averaged across each source language. Full results can be found in Appendix \ref{appendix:full_wmt_results}.}
  \label{table:main_results}
  \centering
  \begin{tabular}{lcc}
    \toprule
    Model                       & $|\tau|$ & $|r|$ \\
    \midrule
    BLEU                        & $0.292$ & $0.423$ \\
    TER                         & $0.352$ & $0.475$ \\
    ROUGE                       & $0.354$ & $0.518$ \\
    METEOR                      & $0.301$ & $0.443$ \\
    chrF++                      & $0.396$ & $0.578$ \\
    BLEURT-large                & $0.625$ & $0.818$ \\
    BERTScore-RoBERTa$_{\small\textrm{large}}$     & $0.567$ & $0.759$ \\
    BERTScore-T5$_{\small\textrm{large}}$          & $0.536$ & $0.738$ \\
    BERTScore-DeBERTa$_{\small\textrm{large}}$     & $0.580$ & $0.773$ \\
    MoverScore                  & $0.322$ & $0.454$ \\
    \midrule
    ParaBLEU$_{\small\textrm{large}}$      & $\mathbf{0.653}$ & $\mathbf{0.843}$ \\
    ParaBLEU$_{\small\textrm{base}}$      & $0.589$ & $0.785$ \\
    \bottomrule
  \end{tabular}
\end{table}
ParaBLEU results on WMT17 are given in Table \ref{table:main_results}, along with a number of baselines. Baselines include BLEURT, described in Section \ref{section:related_work}), along with BERTScore, a non-learned neural metric that uses a matching algorithm on top of neural word embeddings, similar to $n$-gram matching approaches. MoverScore  \cite{zhao2019moverscore} is similar to BERTScore, but uses an optimal transport algorithm. BLEU, ROUGE, METEOR and chrF++ are widely used $n$-gram-based methods, working at the word, subword or character level. TER is an edit-distance-based metric, similar to WER.

ParaBLEU$_{\small\textrm{large}}$ achieves new state-of-the-art results on WMT17, exceeding the previous state-of-the-art approach, BLEURT, on both correlation metrics. We note that non-neural metrics perform the worst, of which the character-level $n$-gram-matching algorithm chrF++ performs the best. Non-learned neural metrics (BERTScore and MoverScore) tend to perform better, and learned neural metrics (BLEURT and ParaBLEU) perform the best. BLEU, the most widely used metric, has the poorest correlation with human judgements. This is consistent with results seen previously in the literature \cite{zhang2019bertscore, sellam2020bleurt}. The significant drop in performance from ParaBLEU$_{\small\textrm{large}}$ to ParaBLEU$_{\small\textrm{base}}$ highlights the benefit of larger, more expressive pretrained language models.

\begin{figure}[t]
  \centering
  \includegraphics[width=0.50\textwidth]{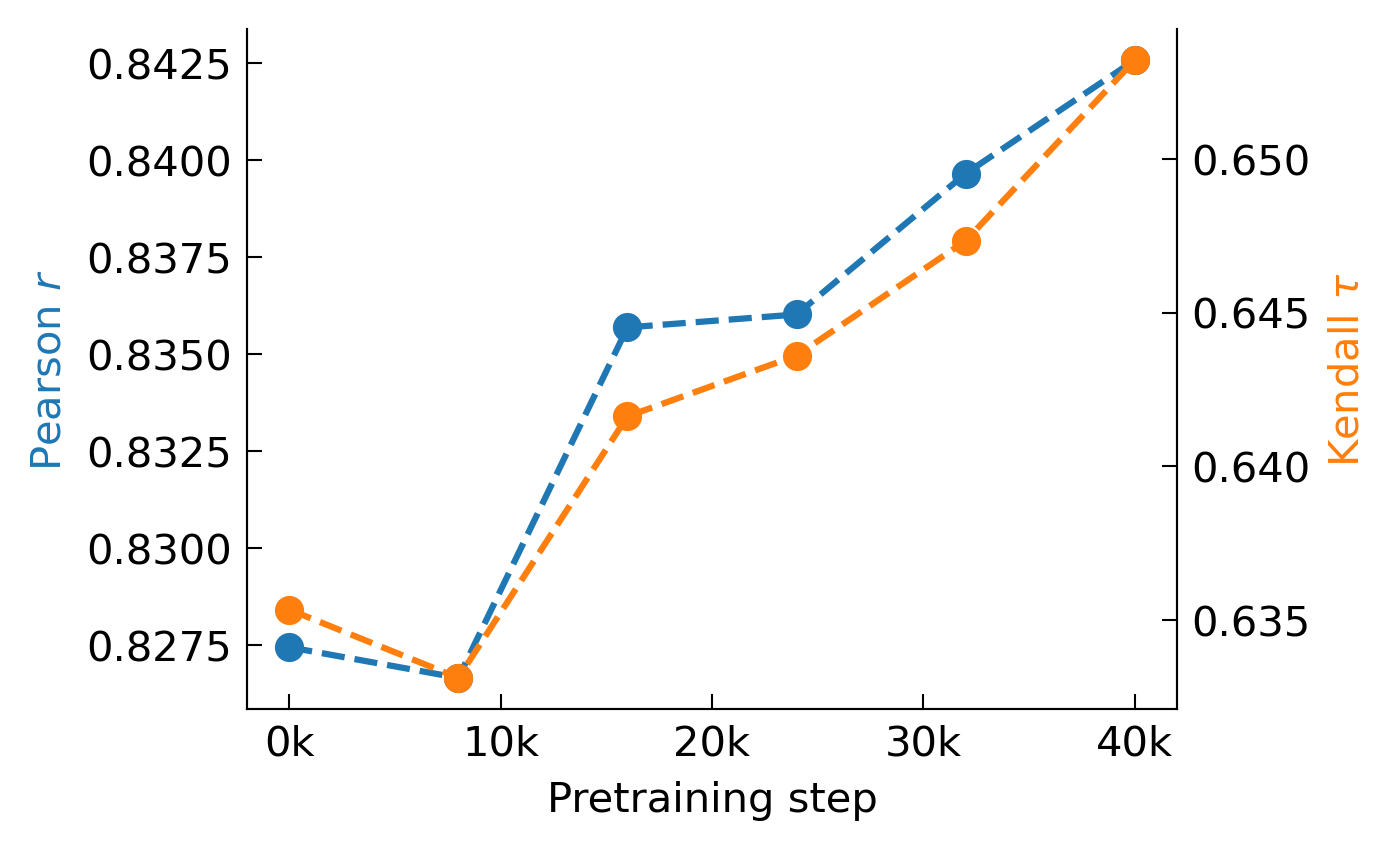}
  \includegraphics[width=0.48\textwidth]{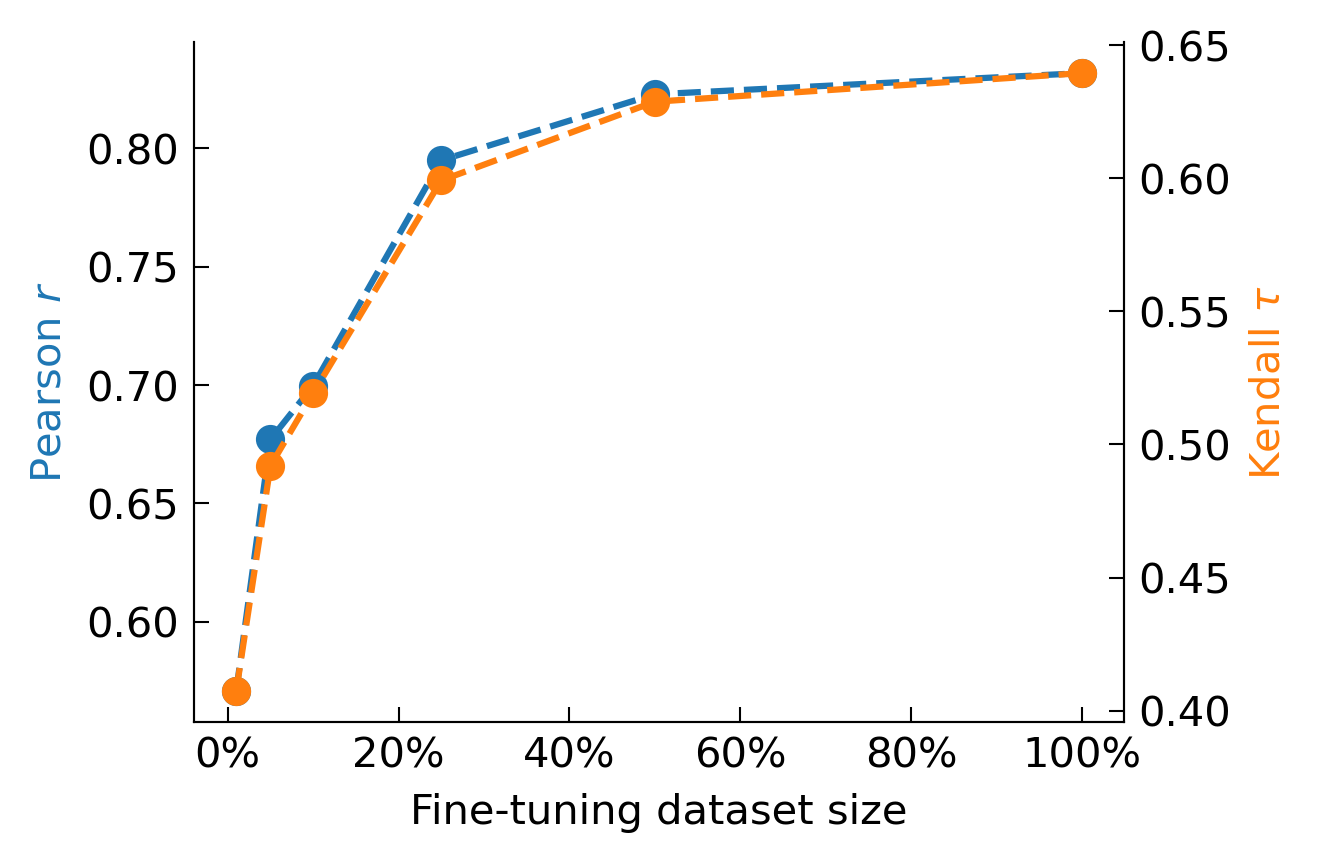}
  \caption{Performance of ParaBLEU$_{\small\textrm{large}}$ on WMT17 as a function of number of pretraining steps (left) and the fine-tuning dataset size (right). Note that the Pearson $r$ results (blue) use the left $y$-axis, whereas Kendall $\tau$ (orange) uses the right $y$-axis.}
  \label{figure:robustness}
\end{figure}

Figure \ref{figure:robustness} probes performance as a function of number of pretraining steps and the size of the fine-tuning dataset for ParaBLEU$_{\small\textrm{large}}$. As expected, pretraining for longer increases downstream task performance. However, we note that $40$k steps, approximately $4$ epochs of ParaCorpus, does not yet reach diminishing returns on WMT17 performance. We therefore recommend pretraining for significantly longer. Both BERT and RoBERTa are pretrained for $40$ epochs \cite{liu2019roberta, lan2019albert}; the T5 authors ablate their dataset size at a fixed number of steps and conclude that performance does not significantly degrade up to and including $64$ epochs \cite{raffel2019exploring}; conversely, the BLEURT authors see diminishing returns on downstream task performance after $2$ pretraining epochs \cite{sellam2020bleurt}.

For the fine-tuning dataset size study, we consistently use a validation set size of $25\%$ to facilitate the small-data results. Despite the training set (the English subsets of WMT15 and WMT16) forming a relatively small dataset, ParaBLEU$_{\small\textrm{large}}$ trained on $50\%$ of the available data ($2,010$ training examples, $670$ validation examples) still beats the previous state-of-the-art, BLEURT, yielding a Pearson correlation of $0.823$. The impact of reducing the train size from $100\%$ ($4,020$ training examples, $1,340$ validation examples) to $25\%$ ($1,005$ training examples, $335$ validation examples) has a relatively small effect on performance, reducing Pearson $r$ from $0.832$ to $0.795$. With a dataset size of only $1\%$ ($40$ training examples, $14$ validation examples), ParaBLEU$_{\small\textrm{large}}$ achieves a Pearson $r$ of $0.571$, still correlating significantly more strongly with human judgements than BLEU, TER, ROUGE, METEOR and MoverScore. We attribute this to the suitability of the generalized pretraining objective for priming the model for paraphrase evaluation tasks.

\subsection{Ablations}\label{section:ablations}
\begin{table}[t]
  \caption{Ablation results on WMT17. The metrics reported are the absolute Kendall $|\tau|$ and Pearson $|r|$ correlation coefficients averaged across each reference language.}
  \label{table:ablations}
  \centering
  \begin{tabular}{lcccc}
    \toprule
    Model                                               & $|\tau|$ & $|r|$ \\
    \midrule
    Baseline (ParaBLEU$_{\small\textrm{large}}$)                   & $\mathbf{0.653}$ & $\mathbf{0.843}$ \\
    \midrule
    No MLM loss ($\mathcal{L}_\textrm{MLM}$)            & $0.633$ & $0.826$ \\
    No autoregressive loss ($\mathcal{L}_\textrm{AR}$)  & $0.642$ & $0.834$ \\
    No entailment classification loss ($\mathcal{L}_\textrm{CLS}$)     & $0.644$ & $0.837$ \\
    \bottomrule
  \end{tabular}
\end{table}

To more directly test the hypotheses in Section \ref{section:inductive_biases}, we perform ablations in which we remove each component of the factorized objective in turn. The results of this are shown in Table \ref{table:ablations}. Each part of the objective is associated with an increase in downstream task performance. The most significant degradation comes from removing the MLM loss. Possible reasons for this include: the MLM loss' contribution to candidate acceptability judgement are crucial; the MLM loss acts as a regularizer, encouraging the edit encoder to represent paraphrases in linguistic concepts rather than low-level edits; and the MLM loss further encourages bitext alignment behaviour, as described in Section \ref{section:inductive_biases}. 

\section{One-shot paraphrase generation}\label{section:generation}
\begin{table}[hp!]
  \small
  \caption{One-shot paraphrase generation results. In every case, the reference is \textit{``Congo's war began in 1998 when Uganda and Rwanda invaded to back rebels fighting to topple the central government.''} The difference between the identity and non-identity version is shown in boldface. `Ent.' is the predicted demonstration reference/candidate entailment probability from the classification loss.}
  \label{table:generation_results}
  \centering
  \begin{adjustbox}{center}
  \bgroup
\def\arraystretch{1.5}
  \begin{tabular}{p{0.08\linewidth}p{0.23\linewidth}p{0.23\linewidth}p{0.05\linewidth}|p{0.30\linewidth}}
    \toprule
     Type & Demo reference & Demo candidate & Ent. & Generated candidate  \\
    \midrule
     negation & Both bidders agreed to assume about $\$90$ million in debt owed on the planes & The bidders did not agree to assume about $\$90$ million in debt owed on the planes & $0.0\%$ & The Congo War began in 1998 when Uganda and Rwanda \textbf{did not invade} to back rebels fighting to topple the central government. \\
     identity (negation) & The bidders did not agree to assume about $\$90$ million in debt owed on the planes & The bidders did not agree to assume about $\$90$ million in debt owed on the planes & $92.6\%$ & The Congo War began in 1998 when Uganda and Rwanda invaded to back rebels fighting to topple the central government. \\
     
     \midrule
     summary & Both bidders agreed to assume about $\$90$ million in debt owed on the planes & They agreed to assume the debt & $99.0\%$ & \textbf{They invaded to back} the rebels fighting to topple the central government. \\
     identity (summary) & They agreed to assume the debt & They agreed to assume the debt & $95.6\%$ & Congo began in 1998 when Uganda and Rwanda invaded to support the rebels fighting to topple the central government. \\
     
     \midrule
     passive & Both bidders agreed to assume about $\$90$ million in debt owed on the planes & $\$90$ million in debt owed on the planes was agreed to be assumed by both bidders & $99.8\%$ & \textbf{The war in Congo} began in 1998 when Uganda and Rwanda \textbf{were invaded to back} rebels fighting to topple the central government. \\
     identity (passive) & $\$90$ million in debt owed on the planes was agreed to be assumed by both bidders & $\$90$ million in debt owed on the planes was agreed to be assumed by both bidders & $92.7\%$ & Congo War began in 1998 when Uganda and Rwanda invaded to back rebels fighting to topple the central government. \\
     
     \midrule
     conditional perfect & Both bidders agreed to assume about $\$90$ million in debt owed on the planes & Both bidders would have agreed to assume about $\$90$ million in debt owed on the planes & $69.4\%$ & \textbf{Congo's war} began in 1998, when Uganda and Rwanda \textbf{would have} invaded to back rebels fighting to topple the central government. \\
     identity (conditional perfect) & Both bidders would have agreed to assume about $\$90$ million in debt owed on the planes & Both bidders would have agreed to assume about $\$90$ million in debt owed on the planes & $90.7\%$ & Congo War began in 1998 when Uganda and Rwanda invaded to back rebels fighting to topple the central government. \\
     
     \midrule
     personal & Both bidders agreed to assume about $\$90$ million in debt owed on the planes & We agreed to assume about $\$90$ million in debt owed on the planes & $95.4\%$ & \textbf{We started the Congo war} in 1998 when Uganda and Rwanda invaded to support rebels fighting to topple the central government. \\
     identity (personal) & We agreed to assume about $\$90$ million in debt owed on the planes & We agreed to assume about $\$90$ million in debt owed on the planes & $93.2\%$ & The Congo War began in 1998 when Uganda and Rwanda invaded to back rebels fighting to topple the central government. \\
     
     \midrule
     informal & Both bidders agreed to assume about $\$90$ million in debt owed on the planes & Both guys agreed to take on like $\$90$ million that they owed for some planes & $98.8\%$ & \textbf{Congo's war started} in 1998 when Uganda and Rwanda invaded to \textbf{get some} rebels fighting to topple the central government. \\
     identity (informal) & Both guys agreed to take on like $\$90$ million that they owed for some planes & Both guys agreed to take on like $\$90$ million that they owed for some planes & $94.5\%$ & Congo War began in 1998 when Uganda and Rwanda invaded to back rebels fighting to topple the central government. \\
    \bottomrule
  \end{tabular}
  \egroup
  \end{adjustbox}

\end{table}

As our final study, we exploit the generative nature of the pretraining architecture to test our claim that the edit encoder reasons in high-level paraphrastic concepts rather than low-level edits. To do this, we diverge from the pretraining setup, in which the same reference text is passed to both the edit encoder and the sequence-to-sequence model, by passing a different, unseen reference to the sequence-to-sequence model. Akin to \cite{brown2020language, gao2020making}, the hope is that the `demonstration paraphrase' acts as a conditioning factor for paraphrasing the unseen sentence in a similar way.

If the model is reasoning in low-level edits or otherwise `cheating', we expect to see:
\begin{itemize}
    \item Thematic/word leakage from the encoder candidate to the generated candidate, caused by the candidate being autoencoded. This is the undesirable behaviour we sought to address using a bottleneck.
    \item Ungrammatical or otherwise unacceptable output with made-up words and/or bad word order, caused by the encoding of low-level edits scrambling the generator reference tokens. 
\end{itemize}
If the model is reasoning in high-level paraphrastic concepts, we expect to see:
\begin{itemize}
    \item Consistently grammatical, acceptable output.
    \item The flavour of the paraphrase mirroring the conditioning, e.g. the altering of a linguistic style, mood or tense.
\end{itemize}

We generate text using beam-search \cite{medress1977speech}. We sample references at random from the MRPC dataset. The demonstration candidate is a hand-crafted paraphrase of the demonstration reference that embodies a pre-specified paraphrase type. We report the predicted entailment score of the demonstration reference and candidate, along with the candidate generated by the model. The examples in Table \ref{table:generation_results} and in Appendix \ref{appendix:more_generation_examples} are selected at random.

The generation results are shown in Table \ref{table:generation_results}. We include two sets of results for each paraphrastic type (e.g. `negative'): one where the demonstration reference/candidate differ in this concept, and one where both embody the concept. Since we wish to encode the \emph{difference} between the demonstration reference/candidate texts, the desired behaviour when the demonstration pair is identical is no change. If this is not the case, it is likely that the edit encoder is just autoencoding the candidate using high-level linguistic concepts, similar to linguistic style transfer. Further examples are given in Appendix \ref{appendix:more_generation_examples}.

The results present a strong case that the encoder is representing high-level paraphrastic concepts. It is able to successfully identify changes in mood, style and tense between the demonstration reference and candidate, and transpose them onto the unseen reference to make a largely grammatical and appropriately paraphrased sentence. We do not see significant leakage of concepts, words or styles between the demonstration candidate and the generated candidate, instead the expected transfer of paraphrase style.

\section{Conclusions}
In this paper, we introduced ParaBLEU, a paraphrase representation learning model and associated paraphrase evaluation metric. We demonstrated that the metric yields state-of-the-art correlation with human paraphrase judgements and is robust to data scarcity. We motivated its pretraining strategy through a set of inductive biases, which we tested through ablation studies. Finally, we reframed the pretraining as a one-shot paraphrase generation model and gathered evidence that ParaBLEU represents meaningful paraphrastic information.

\bibliographystyle{unsrt}
\bibliography{neurips_2021.bib}

\appendix

\section{Pretraining hyperparameters}\label{appendix:model_hyperparameters}
Table \ref{tab:hyperparams} shows the hyperparamters used for the ParaBLEU$_{\small\textrm{base}}$ and ParaBLEU$_{\small\textrm{large}}$ models during pretraining. $\alpha$ and $\beta$ are the loss weights from Equation \ref{eq:loss}.

\begin{table}[H]
    \centering
    \begin{tabular}{l c c}
    \toprule
        Hyperparameter & ParaBLEU$_{\small\textrm{base}}$ & ParaBLEU$_{\small\textrm{large}}$ \\
        \midrule
        Edit encoder base model             & RoBERTa$_{\small\textrm{base}}$  & RoBERTa$_{\small\textrm{large}}$ \\
        Sequence-to-sequence base model     & BART$_{\small\textrm{base}}$     & BART$_{\small\textrm{base}}$ \\
        Batch size (per GPU; examples)      & $64$            & $32$ \\
        Batch size (per GPU; max tokens)    & $16,384$        & $8,192$ \\
        Learning rate (per GPU)             & $4$e-$4$          & $1$e-$4$ \\
        Warmup steps                        & $1,200$         & $2,400$ \\
        Train length (updates)              & $20$k           & $40$k \\
        Train length (epochs)               & $4$             & $4$ \\
        Gradient accumulation steps         & $1$             & $2$ \\
        $\alpha$                            & $2.0$           & $2.0$ \\
        $\beta$                             & $10.0$          & $10.0$  \\
    \bottomrule
    \end{tabular}
    \caption{Pretraining hyperparameters for the ParaBLEU$_{\small\textrm{base}}$ and ParaBLEU$_{\small\textrm{large}}$ models used in this paper. These were adapted for a larger architecture from the RoBERTa paper \cite{liu2019roberta} and not subject to tuning.}
    \label{tab:hyperparams}
\end{table}

\section{Microsoft Research Paraphrase Corpus results}\label{appendix:mrpc_results}

We additionally ran a study on the Microsoft Research Paraphrase Corpus (MRPC) \cite{dolan2005automatically}, a constituent of the GLUE benchmark \cite{wang2018glue}. MRPC contains $5,801$ sentence pairs each accompanied by hand-labelled binary judgement of whether the pair constitutes a paraphrase. The data is split into a train set ($4,076$ sentence pairs of which $2,753$ are paraphrases) and a test set ($1,725$ sentence pairs of which $1,147$ are paraphrases).

We fine-tune our ParaBLEU models on the MRPC train set using the fine-tuning procedure detailed in \cite{liu2019roberta} and predict on the held-out test set. For baselines we use the ALBERT$_{\small\textrm{large}}$ \cite{lan2019albert} and the RoBERTa$_{\small\textrm{large}}$ \cite{liu2019roberta} models fine-tuned using their respective hyperparameters.

\begin{table}[H]
    \centering
    \begin{tabular}{l c c}
    \toprule
        Model & Accuracy & F1 score \\
        \midrule
        ALBERT$_{\small\textrm{large}}$  & $88.2$ & $91.3$ \\        
        RoBERTa$_{\small\textrm{large}}$  & $89.5$ & $92.2$ \\
        \midrule
        ParaBLEU$_{\small\textrm{large}}$  & $88.8$ & $91.5$ \\
        ParaBLEU$_{\small\textrm{base}}$  & $85.2$ & $88.9$ \\
    \bottomrule
    \end{tabular}
    \caption{The results from the Microsoft Research Paraphrase Corpus (MRPC).}
    \label{tab:mrpc_results}
\end{table}

From Table \ref{tab:mrpc_results} we observe that our default model ParaBLEU$_{\small\textrm{large}}$ underperforms compared to the model it is based on, RoBERTa$_{\small\textrm{large}}$. This could be because the hyperparameter sweep we used for our ParaBLEU models (the same sweep as recommended by the authors of RoBERTa$_{\small\textrm{large}}$) is suboptimal and a broader hyperparameter sweep may be required.

\begin{landscape}
\section{Full to-English WMT results}\label{appendix:full_wmt_results}
Table \ref{table:full_wmt_results} shows the full WMT17 results, which are summarized in main paper Table \ref{table:main_results}. See Section \ref{section:results} for more details.

\begin{table}[ht]
  \small
  \caption{Full to-English results for WMT17. The metrics reported are absolute Kendall $|\tau|$ and Pearson $|r|$. Models are fine-tuned on the English subset of WMT15 and WMT16. For a language pair `x-y', the original reference was in language `x', and both human and machine translations are in language `y'. For results averaged across all source languages, see the main paper.}
  \label{table:full_wmt_results}
  \centering
  \begin{tabular}{lccccccccc}
    \toprule
                       & lv-en & tr-en & zh-en & ru-en & de-en & cs-en & fi-en \\
    Model                   & $|\tau|$ / $|r|$ & $|\tau|$ / $|r|$ & $|\tau|$ / $|r|$ & $|\tau|$ / $|r|$ & $|\tau|$ / $|r|$ & $|\tau|$ / $|r|$ & $|\tau|$ / $|r|$ \\
    \midrule
    BLEU                    & 0.215 / 0.334 & 0.313 / 0.461 & 0.344 / 0.488 & 0.313 / 0.431 & 0.259 / 0.372 & 0.255 / 0.373 & 0.342 / 0.503 \\
    TER                     & 0.329 / 0.439 & 0.393 / 0.472 & 0.365 / 0.493 & 0.358 / 0.509 & 0.295 / 0.403 & 0.315 / 0.458 & 0.411 / 0.548 \\
    ROUGE                   & 0.303 / 0.459 & 0.395 / 0.56 & 0.366 / 0.542 & 0.343 / 0.488 & 0.336 / 0.488 & 0.302 / 0.462 & 0.434 / 0.628 \\
    METEOR            & 0.258 / 0.403 & 0.375 / 0.554 & 0.352 / 0.521 & 0.353 / 0.491 & 0.307 / 0.445 & 0.287 / 0.448 & 0.402 / 0.597 \\
    MoverScore              & 0.252 / 0.350 & 0.314 / 0.493 & 0.345 / 0.485 & 0.375 / 0.493 & 0.296 / 0.401 & 0.317 / 0.433 & 0.356 / 0.521 \\
    chrF++                  & 0.333 / 0.520 & 0.432 / 0.614 & 0.405 / 0.593 & 0.415 / 0.588 & 0.365 / 0.534 & 0.35 / 0.523 & 0.475 / 0.678 \\
    BLEURT            & 0.644 / 0.835 & 0.629 / 0.824 & 0.602 / 0.814 & 0.613 / 0.811 & 0.599 / 0.792 & 0.593 / 0.773 & 0.695 / 0.878 \\
    BERTScore-RoBERTa$_{\small\textrm{large}}$ & 0.555 / 0.756 & 0.569 / 0.751 & 0.568 / 0.775 & 0.555 / 0.746 & 0.554 / 0.745 & 0.522 / 0.71 & 0.646 / 0.833 \\
    BERTScore-T5$_{\small\textrm{large}}$      & 0.529 / 0.74 & 0.53 / 0.721 & 0.532 / 0.749 & 0.531 / 0.74 & 0.5 / 0.699 & 0.485 / 0.69 & 0.643 / 0.831 \\
    BERTScore-DeBERTa$_{\small\textrm{large}}$ & 0.581 / 0.785 & 0.579 / 0.755 & 0.584 / 0.795 & 0.576 / 0.771 & 0.561 / 0.751 & 0.537 / 0.729 & 0.642 / 0.825 \\
    \midrule
    ParaBLEU$_{\small\textrm{large}}$  & 0.641 / 0.832 & 0.643 / 0.846 & 0.586 / 0.791 & 0.628 / 0.824 & 0.612 / 0.796 & 0.607 / 0.797 & 0.695 / 0.881 \\
    ParaBLEU$_{\small\textrm{base}}$   & 0.603 / 0.805 & 0.627 / 0.824 & 0.565 / 0.777 & 0.580 / 0.780 & 0.568 / 0.764 & 0.530 / 0.704 & 0.649 / 0.838 \\
    \bottomrule
  \end{tabular}
\end{table}

\pagebreak

\section{More generation examples}\label{appendix:more_generation_examples}
This section includes additional examples of one-shot paraphrase generation sampled from the MRPC dataset. See Section \ref{section:generation} for more information.

\begin{table}[ht]
  \small
  \caption{Additional one-shot paraphrase generation results. `Ent. score' is the predicted demonstration reference/candidate entailment probability from the classification loss. See Section \ref{section:generation} for more information.}
  \centering
  \begin{adjustbox}{center}
  \bgroup
\def\arraystretch{2.0}
  \begin{tabular}{cp{0.08\linewidth}p{0.15\linewidth}p{0.15\linewidth}p{0.05\linewidth}p{0.20\linewidth}|p{0.20\linewidth}}
    \toprule
    \# & Type & Demo reference & Demo candidate & Ent. score & Reference & Generated candidate  \\
    \midrule
     1 & identity & Both bidders agreed to assume about \$ 90 million in debt owed on the planes & Both bidders agreed to assume about \$ 90 million in debt owed on the planes & 90.0\% & Robert Stewart , a spokesman for Park Place , the parent company of Caesars Palace , said he was surprised by the court 's decision . & Robert Stewart, a spokesman for Park Place, the parent company of Caesars Palace, said he was surprised by the court decision. \\
    2 & identity & Both bidders agreed to assume about \$ 90 million in debt owed on the planes & Both bidders agreed to assume about \$ 90 million in debt owed on the planes & 90.0\% & Democrats now hope to increase the value of awards proposed by Hatch and to create a mechanism to ensure the fund remains solvent . & Democrats now hope to increase the value of awards proposed by Hatch and create a mechanism to ensure that the fund remains solvent. \\
    3 & identity & Both bidders agreed to assume about \$ 90 million in debt owed on the planes & Both bidders agreed to assume about \$ 90 million in debt owed on the planes & 90.0\% & Indonesia 's army has often been accused of human rights abuses during GAM 's battle for independence , charges it has generally denied while accusing the separatists of committing rights violations . & Indonesia's army has often been accused of human rights abuses during GAM's battle for independence, charges it generally denied while accusing the separatists of committing rights violations. \\
    4 & identity & Both bidders agreed to assume about \$ 90 million in debt owed on the planes & Both bidders agreed to assume about \$ 90 million in debt owed on the planes & 90.0\% & Progress Software plans to acquire privately held DataDirect Technologies for about \$ 88 million in cash , the companies said Friday . & Progress Software plans to acquire privately held DataDirect Technologies for about \$ 88 million in cash, the companies said Friday. \\
    5 & identity & Both bidders agreed to assume about \$ 90 million in debt owed on the planes & Both bidders agreed to assume about \$ 90 million in debt owed on the planes & 90.0\% & A positive PSA test has to be followed up with a biopsy or other procedures before cancer can be confirmed . & A positive PSA test must be followed up with biopsy or other procedures before cancer can be confirmed. \\
    \bottomrule
  \end{tabular}
  \egroup
  \end{adjustbox}
\end{table}

\begin{table}[ht]
  \small
  \caption{Additional one-shot paraphrase generation results. `Ent. score' is the predicted demonstration reference/candidate entailment probability from the classification loss.}
  \centering
  \begin{adjustbox}{center}
  \bgroup
\def\arraystretch{2.0}
  \begin{tabular}{cp{0.08\linewidth}p{0.15\linewidth}p{0.15\linewidth}p{0.05\linewidth}p{0.20\linewidth}|p{0.20\linewidth}}
    \toprule
    \# & Type & Demo reference & Demo candidate & Ent. score & Reference & Generated candidate  \\
    \midrule
     6 & negation & Both bidders agreed to assume about \$ 90 million in debt owed on the planes & The bidders did not agree to assume about \$ 90 million in debt owed on the planes & 0.0\% & Robert Stewart , a spokesman for Park Place , the parent company of Caesars Palace , said he was surprised by the court 's decision . & Robert Stewart, a spokesman for Park Place, the parent company of Caesars Palace, did not say he was surprised by the court decision. \\
7 & negation & Both bidders agreed to assume about \$ 90 million in debt owed on the planes & The bidders did not agree to assume about \$ 90 million in debt owed on the planes & 0.0\% & Democrats now hope to increase the value of awards proposed by Hatch and to create a mechanism to ensure the fund remains solvent . & Democrats now hope to increase the value of awards proposed by Hatch and to create a mechanism to ensure the fund does not remain solvent. \\
8 & negation & Both bidders agreed to assume about \$ 90 million in debt owed on the planes & The bidders did not agree to assume about \$ 90 million in debt owed on the planes & 0.0\% & Indonesia 's army has often been accused of human rights abuses during GAM 's battle for independence , charges it has generally denied while accusing the separatists of committing rights violations . & The Indonesian army did not often be accused of human rights abuses during the GAM struggle for independence, charges it generally denied while accusing the separatists of committing rights violations. \\
9 & negation & Both bidders agreed to assume about \$ 90 million in debt owed on the planes & The bidders did not agree to assume about \$ 90 million in debt owed on the planes & 0.0\% & Progress Software plans to acquire privately held DataDirect Technologies for about \$ 88 million in cash , the companies said Friday . & Progress Software plans to acquire the privately held DataDirect Technologies for about \$ 88 million in cash, the companies said Friday. \\
10 & negation & Both bidders agreed to assume about \$ 90 million in debt owed on the planes & The bidders did not agree to assume about \$ 90 million in debt owed on the planes & 0.0\% & A positive PSA test has to be followed up with a biopsy or other procedures before cancer can be confirmed . & A positive PSA test does not have to be followed up with biopsy or other procedures before cancer can be confirmed. \\
    \bottomrule
  \end{tabular}
  \egroup
  \end{adjustbox}
\end{table}

\begin{table}[ht]
  \small
  \caption{Additional one-shot paraphrase generation results. `Ent. score' is the predicted demonstration reference/candidate entailment probability from the classification loss.}
  \centering
  \begin{adjustbox}{center}
  \bgroup
\def\arraystretch{2.0}
  \begin{tabular}{cp{0.08\linewidth}p{0.15\linewidth}p{0.15\linewidth}p{0.05\linewidth}p{0.20\linewidth}|p{0.20\linewidth}}
    \toprule
    \# & Type & Demo reference & Demo candidate & Ent. score & Reference & Generated candidate  \\
    \midrule
   11 & summary & Both bidders agreed to assume about \$ 90 million in debt owed on the planes & They agreed to assume the debt & 99.0\% & Robert Stewart , a spokesman for Park Place , the parent company of Caesars Palace , said he was surprised by the court 's decision . & They said they were surprised by the decision. \\
12 & summary & Both bidders agreed to assume about \$ 90 million in debt owed on the planes & They agreed to assume the debt & 99.0\% & Democrats now hope to increase the value of awards proposed by Hatch and to create a mechanism to ensure the fund remains solvent . & They hoped to increase the value of the awards. \\
13 & summary & Both bidders agreed to assume about \$ 90 million in debt owed on the planes & They agreed to assume the debt & 99.0\% & Indonesia 's army has often been accused of human rights abuses during GAM 's battle for independence , charges it has generally denied while accusing the separatists of committing rights violations . & They generally accused the separatists of committing rights violations. \\
14 & summary & Both bidders agreed to assume about \$ 90 million in debt owed on the planes & They agreed to assume the debt & 99.0\% & Progress Software plans to acquire privately held DataDirect Technologies for about \$ 88 million in cash , the companies said Friday . & They wanted to acquire DataDirect Technologies. \\
15 & summary & Both bidders agreed to assume about \$ 90 million in debt owed on the planes & They agreed to assume the debt & 99.0\% & A positive PSA test has to be followed up with a biopsy or other procedures before cancer can be confirmed . & They had to confirm the PSA test. \\
    \bottomrule
  \end{tabular}
  \egroup
  \end{adjustbox}
\end{table}

\begin{table}[ht]
  \small
  \caption{Additional one-shot paraphrase generation results. `Ent. score' is the predicted demonstration reference/candidate entailment probability from the classification loss.}
  \centering
  \begin{adjustbox}{center}
  \bgroup
\def\arraystretch{2.0}
  \begin{tabular}{cp{0.08\linewidth}p{0.15\linewidth}p{0.15\linewidth}p{0.05\linewidth}p{0.20\linewidth}|p{0.20\linewidth}}
    \toprule
    \# & Type & Demo reference & Demo candidate & Ent. score & Reference & Generated candidate  \\
    \midrule
     16 & passive & Both bidders agreed to assume about \$ 90 million in debt owed on the planes & \$ 90 million in debt owed on the planes was agreed to be assumed by both bidders & 99.8\% & Robert Stewart , a spokesman for Park Place , the parent company of Caesars Palace , said he was surprised by the court 's decision . & The court decision was surprised by Robert Stewart, a spokesman for Park Place, the parent company of Caesars Palace. \\
17 & passive & Both bidders agreed to assume about \$ 90 million in debt owed on the planes & \$ 90 million in debt owed on the planes was agreed to be assumed by both bidders & 99.8\% & Democrats now hope to increase the value of awards proposed by Hatch and to create a mechanism to ensure the fund remains solvent . & The value of awards proposed by Hatch is now hoped to increase by Democrats and create a mechanism to ensure the fund remains solvent. \\
18 & passive & Both bidders agreed to assume about \$ 90 million in debt owed on the planes & \$ 90 million in debt owed on the planes was agreed to be assumed by both bidders & 99.8\% & Indonesia 's army has often been accused of human rights abuses during GAM 's battle for independence , charges it has generally denied while accusing the separatists of committing rights violations . & Human rights abuses in the battle for independence was often accused by Indonesia's army, charges which it generally denied while accusing the separatists of committing rights violations. \\
19 & passive & Both bidders agreed to assume about \$ 90 million in debt owed on the planes & \$ 90 million in debt owed on the planes was agreed to be assumed by both bidders & 99.8\% & Progress Software plans to acquire privately held DataDirect Technologies for about \$ 88 million in cash , the companies said Friday . & The privately held DataDirect Technologies plans to be acquired by Progress Software in cash for about \$ 88 million. \\
20 & passive & Both bidders agreed to assume about \$ 90 million in debt owed on the planes & \$ 90 million in debt owed on the planes was agreed to be assumed by both bidders & 99.8\% & A positive PSA test has to be followed up with a biopsy or other procedures before cancer can be confirmed . & A positive PSA test has to be confirmed in biopsy or other procedures before cancer can be followed up. \\
    \bottomrule
  \end{tabular}
  \egroup
  \end{adjustbox}
\end{table}

\begin{table}[ht]
  \small
  \caption{Additional one-shot paraphrase generation results. `Ent. score' is the predicted demonstration reference/candidate entailment probability from the classification loss.}
  \centering
  \begin{adjustbox}{center}
  \bgroup
\def\arraystretch{2.0}
  \begin{tabular}{cp{0.08\linewidth}p{0.15\linewidth}p{0.15\linewidth}p{0.05\linewidth}p{0.20\linewidth}|p{0.20\linewidth}}
    \toprule
    \# & Type & Demo reference & Demo candidate & Ent. score & Reference & Generated candidate  \\
    \midrule
     21 & conditional perfect & Both bidders agreed to assume about \$ 90 million in debt owed on the planes & Both bidders would have agreed to assume about \$ 90 million in debt owed on the planes & 69.4\% & Robert Stewart , a spokesman for Park Place , the parent company of Caesars Palace , said he was surprised by the court 's decision . & Robert Stewart, a spokesman for Caesars Palace, the parent company of Park Place, said he would have been surprised by the court decision. \\
22 & conditional perfect & Both bidders agreed to assume about \$ 90 million in debt owed on the planes & Both bidders would have agreed to assume about \$ 90 million in debt owed on the planes & 69.4\% & Democrats now hope to increase the value of awards proposed by Hatch and to create a mechanism to ensure the fund remains solvent . & Democrats would now hope to increase the value of awards proposed by Hatch and to create a mechanism to ensure that the fund would remain solvent. \\
23 & conditional perfect & Both bidders agreed to assume about \$ 90 million in debt owed on the planes & Both bidders would have agreed to assume about \$ 90 million in debt owed on the planes & 69.4\% & Indonesia 's army has often been accused of human rights abuses during GAM 's battle for independence , charges it has generally denied while accusing the separatists of committing rights violations . & Indonesia's army would have often been accused of human rights abuses during GAM's battle for independence, charges which it generally denied while accusing the separatists of committing rights violations. \\
24 & conditional perfect & Both bidders agreed to assume about \$ 90 million in debt owed on the planes & Both bidders would have agreed to assume about \$ 90 million in debt owed on the planes & 69.4\% & Progress Software plans to acquire privately held DataDirect Technologies for about \$ 88 million in cash , the companies said Friday . & Progress Software plans to acquire privately held DataDirect Technologies for about \$ 88 million in cash, the companies said Friday. \\
25 & conditional perfect & Both bidders agreed to assume about \$ 90 million in debt owed on the planes & Both bidders would have agreed to assume about \$ 90 million in debt owed on the planes & 69.4\% & A positive PSA test has to be followed up with a biopsy or other procedures before cancer can be confirmed . & A positive PSA test would have to be followed up with a biopsy or other procedures before cancer can be confirmed. \\
    \bottomrule
  \end{tabular}
  \egroup
  \end{adjustbox}
\end{table}

\begin{table}[ht]
  \small
  \caption{Additional one-shot paraphrase generation results. `Ent. score' is the predicted demonstration reference/candidate entailment probability from the classification loss.}
  \centering
  \begin{adjustbox}{center}
  \bgroup
\def\arraystretch{2.0}
  \begin{tabular}{cp{0.08\linewidth}p{0.15\linewidth}p{0.15\linewidth}p{0.05\linewidth}p{0.20\linewidth}|p{0.20\linewidth}}
    \toprule
    \# & Type & Demo reference & Demo candidate & Ent. score & Reference & Generated candidate  \\
    \midrule
   26 & unrelated & Both bidders agreed to assume about \$ 90 million in debt owed on the planes & Bush turned out a statement yesterday thanking the commission for its work , and said , " Our journey into space will go on . " & 0.2\% & Robert Stewart , a spokesman for Park Place , the parent company of Caesars Palace , said he was surprised by the court 's decision . & A spokesman for Caesars Palace, the parent company of Park Place, said he was surprised by the court's decision. \\
27 & unrelated & Both bidders agreed to assume about \$ 90 million in debt owed on the planes & Bush turned out a statement yesterday thanking the commission for its work , and said , " Our journey into space will go on . " & 0.2\% & Democrats now hope to increase the value of awards proposed by Hatch and to create a mechanism to ensure the fund remains solvent . & Hatch now hopes to raise the value of awards proposed by Democrats, and create a mechanism to ensure that the fund remains solvent. \\
28 & unrelated & Both bidders agreed to assume about \$ 90 million in debt owed on the planes & Bush turned out a statement yesterday thanking the commission for its work , and said , " Our journey into space will go on . " & 0.2\% & Indonesia 's army has often been accused of human rights abuses during GAM 's battle for independence , charges it has generally denied while accusing the separatists of committing rights violations . & During the battle for independence, GAM's army has often been accused of human rights abuses, which it generally denied, while accusing the separatists of committing rights violations. \\
29 & unrelated & Both bidders agreed to assume about \$ 90 million in debt owed on the planes & Bush turned out a statement yesterday thanking the commission for its work , and said , " Our journey into space will go on . " & 0.2\% & Progress Software plans to acquire privately held DataDirect Technologies for about \$ 88 million in cash , the companies said Friday . & The company announced on Friday that Progress Software plans to acquire DataDirect Technologies, for about \$ 88 million in cash. \\
30 & unrelated & Both bidders agreed to assume about \$ 90 million in debt owed on the planes & Bush turned out a statement yesterday thanking the commission for its work , and said , " Our journey into space will go on . " & 0.2\% & A positive PSA test has to be followed up with a biopsy or other procedures before cancer can be confirmed . & The report states that, before cancer can be confirmed, a positive PSA test must be carried out with a biopsy or other procedures. \\
    \bottomrule
  \end{tabular}
  \egroup
  \end{adjustbox}
\end{table}

\begin{table}[ht]
  \small
  \caption{Additional one-shot paraphrase generation results. `Ent. score' is the predicted demonstration reference/candidate entailment probability from the classification loss.}
  \centering
  \begin{adjustbox}{center}
  \bgroup
\def\arraystretch{2.0}
  \begin{tabular}{cp{0.08\linewidth}p{0.15\linewidth}p{0.15\linewidth}p{0.05\linewidth}p{0.20\linewidth}|p{0.20\linewidth}}
    \toprule
    \# & Type & Demo reference & Demo candidate & Ent. score & Reference & Generated candidate  \\
    \midrule
     31 & personal & Both bidders agreed to assume about \$ 90 million in debt owed on the planes & We agreed to assume about \$ 90 million in debt owed on the planes & 95.4\% & Robert Stewart , a spokesman for Park Place , the parent company of Caesars Palace , said he was surprised by the court 's decision . & Robert Stewart said he was surprised by the court decision. \\
32 & personal & Both bidders agreed to assume about \$ 90 million in debt owed on the planes & We agreed to assume about \$ 90 million in debt owed on the planes & 95.4\% & Democrats now hope to increase the value of awards proposed by Hatch and to create a mechanism to ensure the fund remains solvent . & Democrats now hope to increase the value of awards proposed by Hatch. \\
33 & personal & Both bidders agreed to assume about \$ 90 million in debt owed on the planes & We agreed to assume about \$ 90 million in debt owed on the planes & 95.4\% & Indonesia 's army has often been accused of human rights abuses during GAM 's battle for independence , charges it has generally denied while accusing the separatists of committing rights violations . & Indonesia has often been accused of human rights abuses during the battle for independence. \\
34 & personal & Both bidders agreed to assume about \$ 90 million in debt owed on the planes & We agreed to assume about \$ 90 million in debt owed on the planes & 95.4\% & Progress Software plans to acquire privately held DataDirect Technologies for about \$ 88 million in cash , the companies said Friday . & We said we planned to acquire DataDirect Technologies privately for about \$ 88 million in cash. \\
35 & personal & Both bidders agreed to assume about \$ 90 million in debt owed on the planes & We agreed to assume about \$ 90 million in debt owed on the planes & 95.4\% & A positive PSA test has to be followed up with a biopsy or other procedures before cancer can be confirmed . & We have to follow a positive PSA test with biopsy or other procedures before cancer can be confirmed. \\
    \bottomrule
  \end{tabular}
  \egroup
  \end{adjustbox}
\end{table}

\begin{table}[ht]
  \small
  \caption{Additional one-shot paraphrase generation results. `Ent. score' is the predicted demonstration reference/candidate entailment probability from the classification loss.}
  \centering
  \begin{adjustbox}{center}
  \bgroup
\def\arraystretch{2.0}
  \begin{tabular}{cp{0.08\linewidth}p{0.15\linewidth}p{0.15\linewidth}p{0.05\linewidth}p{0.20\linewidth}|p{0.20\linewidth}}
    \toprule
    \# & Type & Demo reference & Demo candidate & Ent. score & Reference & Generated candidate  \\
    \midrule
  36 & informal & Both bidders agreed to assume about \$ 90 million in debt owed on the planes & Both guys agreed to take on like \$ 90 million that they owed for some planes & 98.8\% & Robert Stewart , a spokesman for Park Place , the parent company of Caesars Palace , said he was surprised by the court 's decision . & Robert Stewart, a spokesman for Park Place, the parent company of Caesars Palace, said they were surprised by the court decision. \\
37 & informal & Both bidders agreed to assume about \$ 90 million in debt owed on the planes & Both guys agreed to take on like \$ 90 million that they owed for some planes & 98.8\% & Democrats now hope to increase the value of awards proposed by Hatch and to create a mechanism to ensure the fund remains solvent . & Democrats now hope to get the value of awards proposed by Hatch and create a mechanism to keep the fund solvent. \\
38 & informal & Both bidders agreed to assume about \$ 90 million in debt owed on the planes & Both guys agreed to take on like \$ 90 million that they owed for some planes & 98.8\% & Indonesia 's army has often been accused of human rights abuses during GAM 's battle for independence , charges it has generally denied while accusing the separatists of committing rights violations . & Indonesia's army often got accused of human rights abuses at GAM's battle for independence, which they generally denied while accusing the separatists of committing rights violations. \\
39 & informal & Both bidders agreed to assume about \$ 90 million in debt owed on the planes & Both guys agreed to take on like \$ 90 million that they owed for some planes & 98.8\% & Progress Software plans to acquire privately held DataDirect Technologies for about \$ 88 million in cash , the companies said Friday . & They wanted to buy some privately held DataDirect Technologies for about \$ 88 million in cash on Friday. \\
40 & informal & Both bidders agreed to assume about \$ 90 million in debt owed on the planes & Both guys agreed to take on like \$ 90 million that they owed for some planes & 98.8\% & A positive PSA test has to be followed up with a biopsy or other procedures before cancer can be confirmed . & Some PSA tests need to be followed up with a biopsy or other procedures before they get cancer confirmed. \\
    \bottomrule
  \end{tabular}
  \egroup
  \end{adjustbox}
\end{table}

\end{landscape}

\end{document}